\documentclass[conference]{IEEEtran}
\IEEEoverridecommandlockouts
\usepackage{cite}
\usepackage{amsmath,amssymb,amsfonts}
\usepackage{algorithm}
\usepackage[noend]{algpseudocode}
\usepackage{placeins}
\usepackage{graphicx}
\usepackage{textcomp}
\usepackage{multicol}
\def\BibTeX{{\rm B\kern-.05em{\sc i\kern-.025em b}\kern-.08em
    T\kern-.1667em\lower.7ex\hbox{E}\kern-.125emX}}
\begin{document}

\title{A Decentralized Mobile Computing Network for Multi-Robot Systems Operations\\
%{\footnotesize \textsuperscript{*}Note: Sub-titles are not captured in Xplore and should not be used}
\thanks{This work was supported by a MOE-Tier 1 grant \#T1MOE17001. JLK is supported by a Presidential Graduate Fellowship from the Singapore MOE.}
}

\author{\IEEEauthorblockN{Jabez Leong Kit\textsuperscript{1}, David Mateo\textsuperscript{2}, Roland Bouffanais\textsuperscript{3}}
\IEEEauthorblockA{\textit{Engineering Product Development} \\
\textit{Singapore University of Technology and Design}\\
Singapore, Singapore \\
\textsuperscript{1}jabez\_leong@mymail.sutd.edu.sg,\textsuperscript{2}david\_mateo@sutd.edu.sg,\textsuperscript{3}bouffanais@sutd.edu.sg}
}

\maketitle
\begin{abstract}
Collective animal behaviors are paradigmatic examples of fully decentralized operations involving complex collective computations such as collective turns in flocks of birds or collective harvesting by ants. These systems offer a unique source of inspiration for the development of fault-tolerant and self-healing multi-robot systems capable of operating in dynamic environments. Specifically, swarm robotics emerged and is significantly growing on these premises. However, to date, most swarm robotics systems reported in the literature involve basic computational tasks---averages and other algebraic operations. In this paper, we introduce a novel Collective computing framework based on the swarming paradigm, which exhibits the key innate features of swarms: robustness, scalability and flexibility. Unlike Edge computing, the proposed Collective computing framework is truly decentralized and does not require user intervention or additional servers to sustain its operations. This Collective computing framework is applied to the complex task of collective mapping, in which multiple robots aim at cooperatively map a large area. Our results confirm the effectiveness of the cooperative strategy, its robustness to the loss of multiple units, as well as its scalability. Furthermore, the topology of the interconnecting network is found to greatly influence the performance of the collective action.

\end{abstract}


\section{Introduction}

Many biological systems are known to be capable of performing highly complex computations in a fully decentralized fashion: e.g. the brain, some insect colonies, aggregating amoeboid cells, etc~\cite{bouffanais15:_desig_contr_swarm_dynam}. For instance, collective turns in flock of birds is one such kind of collective decision making achieved through a fully decentralized computational process. Specifically, each bird in a flock detects the direction of travel of some local neighboring conspecifics (according to a given interaction distance). This flow of behavioral information reaching each bird is then processed by the bird's central nervous system according to specific innate behavioral rules that involve some forms of computation of the available data. This complex process can be better understood when considering basic models of flocking based on self-propelled particles (SPPs). In Vicsek's model for instance, the direction of travel of all agents located within a given radius---a metric interaction distance---is simply averaged---the computational task---by each agent~\cite{vicsek1995novel}. It is important to appreciate the fact that flocking birds (and SPPs) are essentially networked units, and that the topology of the underlying network of interaction plays a pivotal role in the effectiveness of the collective behavior~\cite{komareji13:_resil_contr_dynam_collec_behav}.

Over the past five years, we have been witnessing dramatic advances in sensors, digital signal processing capabilities, low-cost single-board computers, storage devices, low-power communication devices. Simultaneously, the cost of hardware has been following an ever-decreasing trend. Combined with unabated developments in robotic software (the Robot Operating System, ROS, is celebrating its 10th anniversary this year), all these rapid technological advances are revolutionizing our ability to build massively distributed, rapidly deployable, self-calibrating multi-robot systems. This paves the way for a fundamental paradigm shift in robotics, in which large, costly, and task-specific robots are replaced by swarms of small, low-cost, and versatile units~\cite{schultz2013multi}.

Swarm robotic systems allow unsophisticated, low-cost, modular robotic platforms to be dynamically reconfigured into a group capable of achieving a range of effective and responsive cooperative actions well beyond the capabilities of the individuals~\cite{brambilla2013swarm,chamanbaz17:_swarm_enabl_techn_multi_robot_system}. This so-called ``power of masses" requires the constituting units to sense and interact with the environment, while also sharing the sensed data with the rest of the swarm, thereby enabling a very effective form of collective computation. This paradigm of decentralized operations, inspired from natural swarms, offers the possibility of performing global collective computations under a wide range of group sizes (scalability), despite the possible sudden loss of multiple agents (robustness), and under unknown and dynamic circumstances (flexibility)~\cite{brambilla2013swarm}. Scalability, flexibility, and robustness are indubitably appealing features as they open the door to the possibility of operating over very large scales, in fully unstructured and dynamic environments, with progressive and graceful degradation of the system’s operations in the presence of adverse conditions. However, it is critical to acknowledge the fact that the achievement of complex cooperative operations by swarm robotics systems hinges on our ability to tap into the effectiveness of their collective computation capabilities---well beyond the trivial averaging process performed during collective motion.

%\textbf{- Why swarm computing is the next step in robotic system/network.}

%In recent years, research in multi-agent systems has experienced a tremendous growth fueled by notable theoretical developments, but also by the emergence of advanced, large-scale, and complex physical embodiments in the form of decentralized multi-robot and swarming systems~\cite{schultz2013multi}. One key challenge in designing such decentralized multi-robot systems (MRS) is ensuring decentralization at all levels of operation: individual control, sensing, inter-agent communication, and collective decision-making. Swarming systems are a perfect incarnation of these challenges, but also of some highly desirable features such as robustness, scalability and flexibility~\cite{brambilla2013swarm,bouffanais15:_desig_contr_swarm_dynam}. However, these sought-after properties proscribe the use of centralized computing resources, although computations are required by virtually every single system component.

 From a network perspective, collective computing is a paradigm sought after when the classical centralized computing paradigm (with a master node and slaves, i.e. the star network) of parallel computing breaks down~\cite{marjovi2012robotic}. That explains that the research community is actively exploring moving away from Cloud computing, and exploring Edge computing capabilities for certain applications that include mobile robotics. However, with dynamic networks of mobile sensors, robots or vehicles, Edge computing demands to leverage resources---computers, sensors, actuators---that may not always be available or connected to the network. Edge computing still requires a few servers throughout the network to serve the sensor clusters, thereby offering only a partial decentralization of the system. Therefore, Collective computing is one step ahead of the highly sought after Edge computing, bringing about complex computations all the way to the end nodes of the network. This significantly reduces communication resources required by Cloud computing to perform most complex computational tasks. More importantly, Collective computing yields an inherently robust, scalable and flexible computing paradigm since these critical features are natural by-products of the underpinning decentralized swarming framework.

In this paper, we introduce the concept of collective computing and present, in detail, one particular embodiment meant to characterize and illustrate the potential of this approach in terms of scalability and robustness. The considered application is collective mapping by a swarm of networked robots, with varying connectivity rules---i.e. varying the topology of the interconnecting network, and with imposed failure of up to 75\% of the nodes of the system.

\section{Collective Computing and Operation}
% {\bf Problem describing information/decision propagating. Show a figure of how neighbors interact and how these information are used in Local Computation and influencing the Swarm Control Strategy.}
\begin{figure}[!htbp]
\centering
\includegraphics[width=0.3\textwidth]{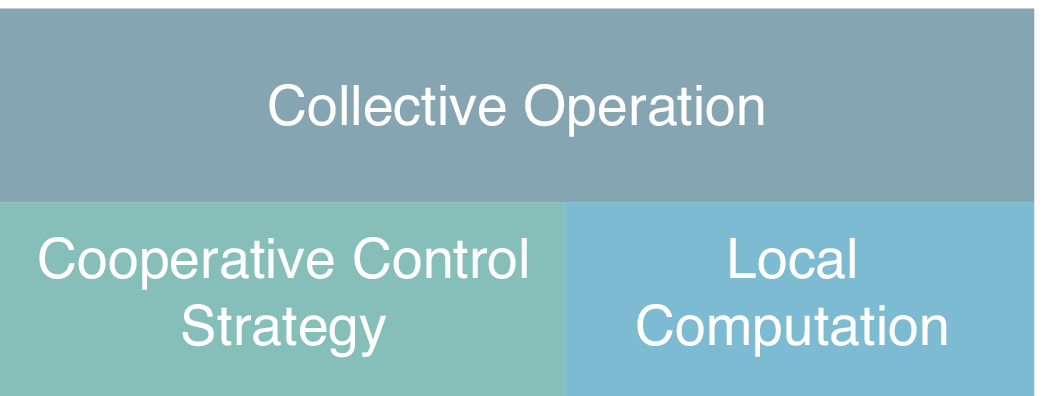}
\caption{\textsf{Schematic of the Collective Operation and Computation framework. Although the cooperative control strategy and the local computation are shown here, for clarity, as two separate boxes, they happen to be deeply intertwined as a consequence of its distribution throughout the interconnecting network.}}
\label{fig:Framework}
\end{figure}

Swarming is known to be a powerful paradigm to achieve a scalable, robust, and flexible \emph{collective operation} performed at spatial (and possibly temporal) scales much larger than the characteristic scales at which single agents operate.
Both natural and artificial swarming systems typically operate by combining a decentralized \emph{cooperative control strategy} with a set of \emph{local computations} that each agent performs using only information from a certain local neighborhood.

\begin{figure}[!htbp]
\centering
\includegraphics[width=2.6in]{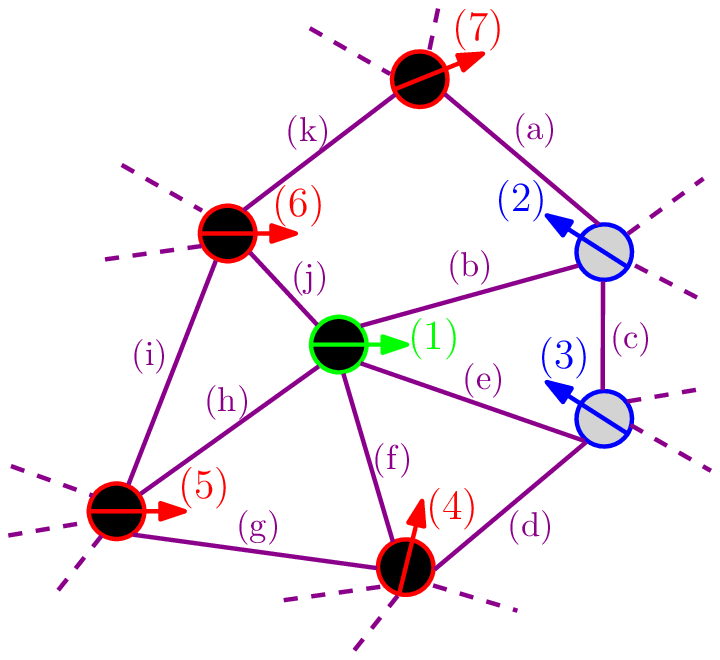}
\caption{\textsf{Schematic of a subset of collectively moving agents within a swarm. Arrows show the direction of travel and straight lines represent the existence of an interaction link between any two agents. Agent (1) receives behavioral information from neighboring agents (2)--(7). If one assumes that agents (2) and (3) are picking up an external signal---e.g., oncoming obstacle---and are responding to it thereby triggering meaningful behavioral information in the form of changes to the agent’s heading represented by blue arrows. This meaningful information reaches agent (1) directly through network edges (b) and (e). This direct signal is reinforced by means of positive feedback loops such as (d)-(f), (a)-(k)-(j), (d)-(g)-(h) and many others. At the same time, agent (1) receives behavioral information from the bulk of the swarm represented by red-colored agents such as (4), (5), (6), and (7) for instance.}}
\label{fig:interaction}
\end{figure}

For instance, flocks of birds are able to keep a consistent but flexible formation during long journeys without the need for a central controller.
This behavior can be modeled by a collective operation with a cooperative control strategy based on global alignment of the birds' direction of motion.
To obtain this global alignment, each agent only has to compute the average direction of motion for its local neighborhood, say its $k$ closest neighbors, and the global, collective movement emerges.
% Instead of solving a large complex computation problem at the system level, these computation are broken down into smaller and simpler Local computation problem solved at the node level. Instead of taking into account every information available to every individual agents, Local Computation only takes into account the information gathered by its neighbors. These Local Computation computes a decision that the individual agents will make in the next iteration, thus influencing its Swarm Control Strategy.
The concept of combining local computation with cooperative control strategy can be illustrated with an example of how a swarm determine its direction of motion based on \textit{Vicsek's model}~\cite{vicsek1995novel}.
For instance, in \cite{komareji13:_resil_contr_dynam_collec_behav}, the local computation of~\eqref{eq:localcomputation} of the swarm shown computes the direction of motion of agent $i$ at time $t+\Delta t$ with shared inputs from its neighbors (see Fig.~\ref{fig:interaction}, by finding the average of the difference of its neighbors and its own direction of motion of time $t$. The updated direction of motion will then be shared with its $k$ nearest neighbors through a network shown in Fig.~\ref{fig:interaction} for the on-going local computations.
It is worth highlighting that the topology of the network is embedded in the definition of the local update rule~\eqref{eq1} and is therefore an integral part of the cooperative control strategy. The properties of this network directly influence the local computational task involved but not its fundamental nature---e.g. for SPPs the computational task is an averaging but over more or less neighbors depending on the local density of agents within the radius of interaction. Moreover,
The local computation outcome affects the dynamics of the node which affects the $k$ nearest neighbor network and those of the swarming units connected to it through a complex propagation process. Both the dynamics of the node and the network defines the cooperative control strategy. It appears clearly that a swarming system operates with local computation and cooperative control strategy.

\begin{equation}\label{eq:localcomputation}
\begin{split}
\theta_i(t+\Delta t) = \theta_i(t) + \frac{\Delta t}{k}[(\theta_j(t)-\theta_i(t))\\
+\dots+(\theta_{j+k+1}(t)-\theta_i(t))]\\
+\eta\xi_i(t)
\end{split}
\end{equation}
where $\eta\xi_i(t)$ is a Gaussian white noise of magnitude $\eta$ since $\xi_i(t) \in[-\pi,\pi]$.

This framework provides a useful guide to review existing robotic functions as collective operations and consider not only the local operations but how different models of cooperative control affect the efficiency of the collective operation. Figure~\ref{fig:Framework} shows the modularity of the framework, yet hides the actual entanglement of both processes due to its distributed nature throughout the interconnecting network underpinning the process.

% \subsection{Swarm Control Strategy}
The effectiveness of the cooperative control strategy depends on (i) the dynamics of the agents, and (ii) the interaction between them.
In network-theoretic terms, this means that the collective operation will be determined by both the dynamics \emph{on} the network and the dynamics \emph{of} the network itself.

In the next section, we present an application of the introduced Collective computing framework for applications of collective mapping by a swarm of networked mobile robots with particular attention paid to scalability and robustness.

% \section{Environmental Sensing}
% {\bf Is this going to outlook?}
% \subsection{General Overview}
% \subsection{Framework}
% \subsection{Implementation}
% \subsection{Experiment and Results}

\subsection{Collective Mapping}
% \subsection{General Overview}
Simultaneous Localisation and Mapping (SLAM) is a complex problem that has received sizable attention from the research community.
This attention has produced some elaborate and efficient solutions for the problem in the case of single-robot SLAM. Recently, these results have been expanded to the multi-robot case in a framework that is referred to as Collaborative SLAM (CoSLAM)~\cite{saeedi2011multiple}.

There are essentially two approaches in the CoSLAM research.
One is to propose new SLAM methods to handle multiple noisy sensors and improve the data association~\cite{gouveia2015computation}---i.e., filtering~\cite{howard2006multi}, scan matching~\cite{diosi2005laser}, and map-merging~\cite{birk2006merging} techniques.
The other is to focus on new multi-robot architectures with efficient data structures~\cite{agarwal2013robust} and distributed computing robotic clusters~\cite{gouveia2015computation,marjovi2012robotic}.
In order to apply these potential solutions to a wide range of environments, attention must be paid to assure that they guarantee not only the robustness but also the scalability of the approach.

Current CoSLAM solutions require a sizable amount of sensor data processing from a central server, that sometimes is performed {\em a posteriori}, i.e. after the robots have completed their explorations.
This requires reliable communication with a central controller, thereby severely limiting the scalability and applicability of the method.
To overcome this serious limitation, one needs a decentralized approach.

In this section, we present the application of the Collective computing framework introduced in the previous section to the task of collective mapping, with simulations of swarming robots allowing us to analyze and assess its effectiveness.
We intend to continue the testing of collective mapping with our swarm of ground vehicles~\cite{chamanbaz17:_swarm_enabl_techn_multi_robot_system}, which has been designed with the ability impose any kind of network topology~\cite{mateo2018optimal}. Therefore, the concepts presented here have been considered with real-life robotic implementations and applications in mind.

\subsection{Simulation}
% In order to showcase the effect of swarming in collective mapping, the SLAM problem is reduced to a mapping problem with the following justifications.
We assume perfect sensory data and localization from the simulated robots.
% In \cite{mateo2018optimal}, the field experiments conducted to show the optimal network topology is performed on a platform developed by our research group. The platform is equipped with six infrared rangefinders, an inertial measurement unit (IMU), two wheel encoders, two light sensors, and XBee communication unit. This platform will be used for future Collective Mapping field experiments.
The local map of each individual robot is classically built with an occupancy grid approach~\cite{elfes1990occupancy} based on their simulated infra-red (IR) sensors.
Each grid cell contains the probability of finding an obstacle. The collective map, defined as the union of all the robots' local maps, is assembled and built to study the efficiency of the approach. However, it is important to stress that it is not needed and that it plays no role in determining the robots' behavior.
The simulated ``sensor range'' is a tile of $6\times 7$ grid cells around it.

To perform the collective mapping operation, each robot computes its own local map based on the environmental data gathered from its own sensors and its neighbors' sensors, with the concept of ``neighbor" understood in its network sense, i.e. the units directly connected to a robot through the underlying network of interaction (see Fig.~\ref{fig:interaction}).
With this updated map, each robot decides independently its next target location by means of a frontier exploration algorithm~\cite{yamauchi1997frontier}.
While this process does not explicitly take into account the position of the neighboring robots, the information gathered from them does affect the map and thus the target location.
Lastly, the robot runs an A$^*$ path planning algorithm~\cite{hart1968formal} in order to reach the target location.
The full process is summarized in Algorithm~\ref{alg:cm}.

\begin{algorithm}
\caption{Collective Mapping algorithm}\label{alg:cm}
\begin{algorithmic}[1]
\Procedure{CollectiveMapping}{$r,R,k,s,map$}%\Comment{The g.c.d. of a and b}
\State $map\gets UpdateOccupancyGrid(map,s)$
\State $neighbors \gets KNearestNeighbors(r,R,k)$
\For{$n$ in $neighbors$}
\State $map\gets UpdateOccupancyGrid(map,s_{n})$
\EndFor
\State $target \gets FrontierExploration(map,r)$
\State $path \gets PathPlanning(r, target)$
\State \textbf{return} $map, path$
\EndProcedure
\end{algorithmic}
\end{algorithm}

The interaction network that determines which robot transmits data to which, is defined using a $k$-nearest neighbor scheme, meaning that at any particular time-step a robot uses the sensing information from its closest $k$ robots.
This means that the network is directed (agent $i$ may be using information from $j$ without $j$ using that of $i$), and dynamic (the network depends on the position of the robots, and is thus affected by their movement). Moreover, it is worth adding that with the particular choice of the $k$-nearest neighbors as interacting units, the network has a spatial embedding~\cite{bouffanais15:_desig_contr_swarm_dynam}.

\begin{figure}[htp]%!htbp
\centering
\includegraphics[width=0.5\textwidth]{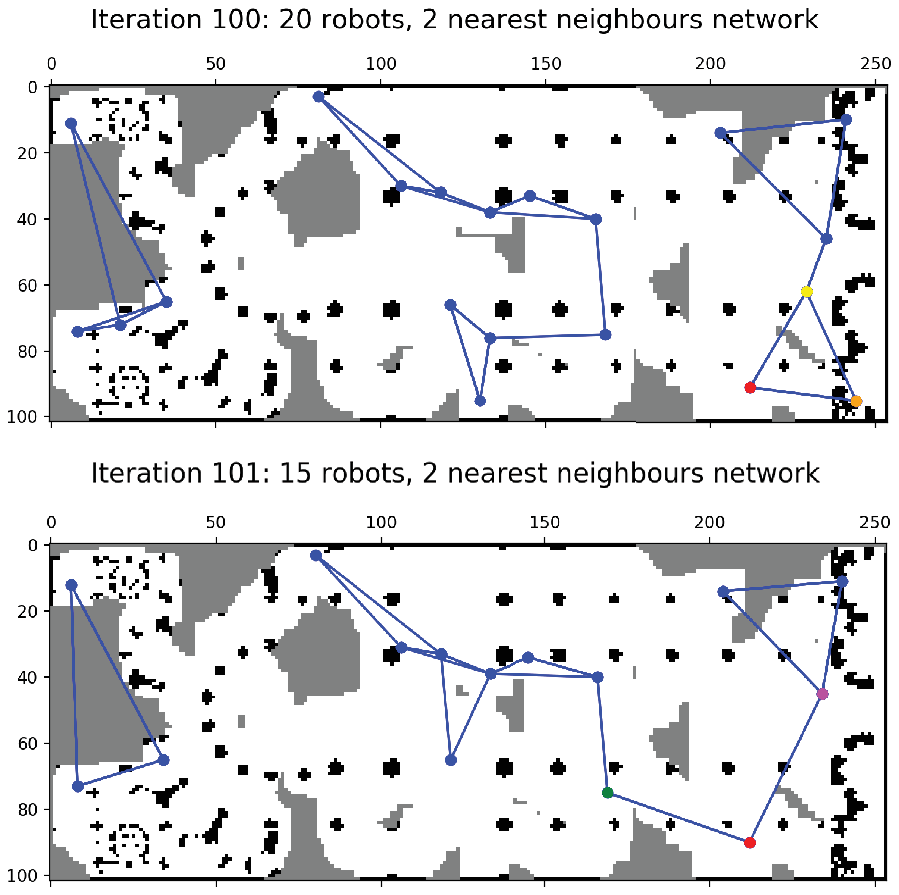}
\caption{\textsf{Two snapshots of the dynamic network obtained during the simulations.
While at any given iterations the system is likely split in disconnected clusters, this instantaneous clusters do communicate with each other thanks to the switching network.
In this example, Robot 14 (red) changes its neighborhood from Robot 15 (yellow) and Robot 19 (orange) to Robot 11 (magenta) and Robot 13 (green).}}
\label{fig:ds2}
\end{figure}

This scheme has been proven in~\cite{mateo2018optimal} to be key in reproducing the collective behavior of natural swarming systems, and to provide surprisingly robust and responsive behaviors with a minimal amount of connections.
For low values of $k$, the \emph{instantaneous} networks generated by this scheme are typically disconnected (see Fig.~\ref{fig:ds2}), but the dynamic stitching of the neighbors allows for the system to be connected over time as the system dynamically evolve over the large area to map.

% \begin{algorithm}
% \caption{Nearest neighbor algorithm}\label{alg:nn}
% \begin{algorithmic}[1]
% \Procedure{NN}{$r,R,num,range$}%\Comment{The g.c.d. of a and b}
% \For{$robot$ in $R$}
% \State $d\gets robot-r$
% \EndFor\label{NNfor}
% \State $nr\gets r_{d\_min}$
% \State \textbf{return} $nr$
% \EndProcedure
% \end{algorithmic}
% \end{algorithm}

The robots only share their current sensed data, meaning that when a robot connects to another it does not receive the history of measurements nor the current local map of the neighbor: this process is Markovian.
Formally, the local map of robot $i$ at time $t$ is obtained by computing the posterior probability $p(m|S_t^i)$ for a collection of sensory data such that

\begin{equation}\label{eq1}
    S_t^i = \bigcup_{t'=1}^t\{s_j(t') ; j| a_{ij}(t')=1\}
\end{equation}
with $a_{ii}(t)=1\, \forall t$.

As this work focuses on the swarming aspect of collective mapping, we have intentionally simplified the (local) mapping technique with a set of assumptions.
Local map-merging can be considered when the robots do not know the relative positions of other robots. As the individual robot has limited processing power, we can distribute the heavy computational work, and thus propose a new strategy that requires less resources.

\subsection{Results}
\begin{figure}[!htbp]
\centering
\includegraphics[width=0.48\textwidth]{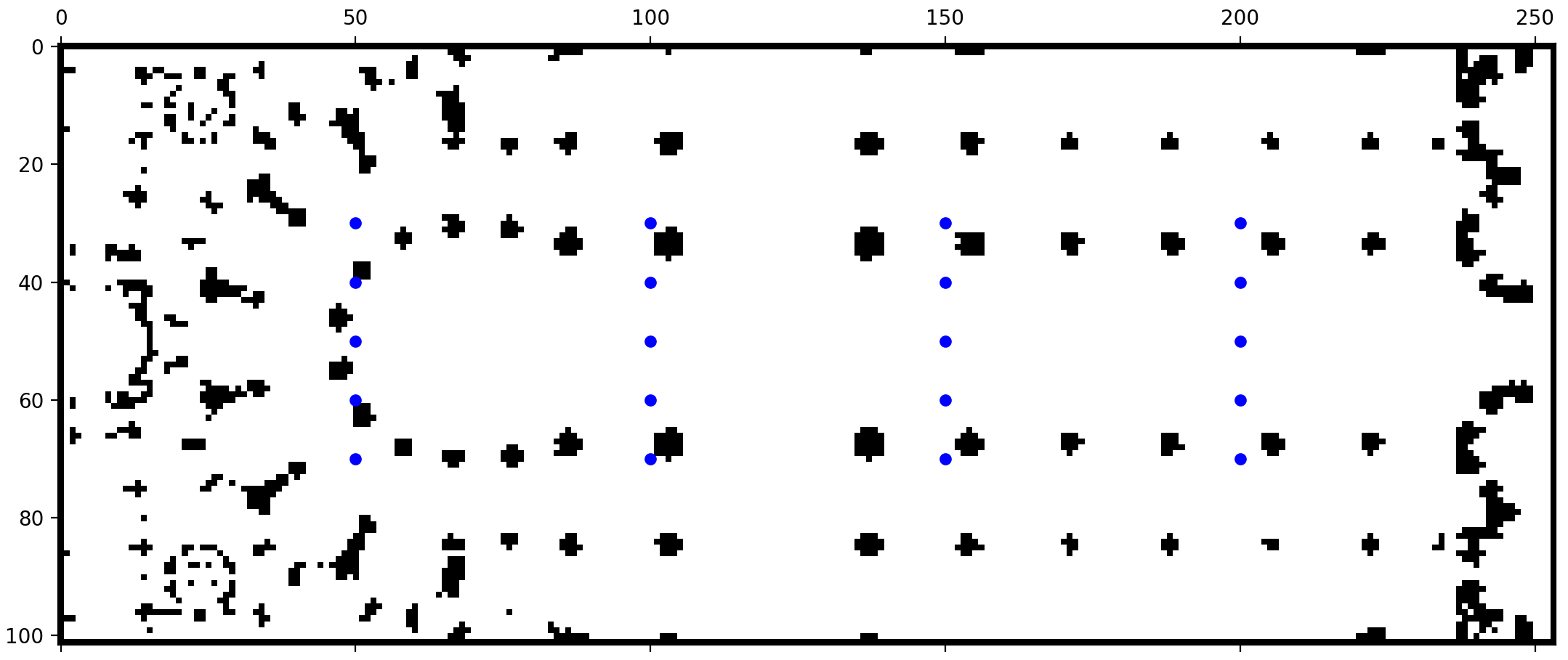}
\caption{\textsf{An interior map of a Basilica used for the Collective Mapping operation. The blue dots represent the initial positions of the simulated robots.}}
\label{church}
\end{figure}

\begin{figure}[!htbp]
\centering
\includegraphics[width=0.48\textwidth]{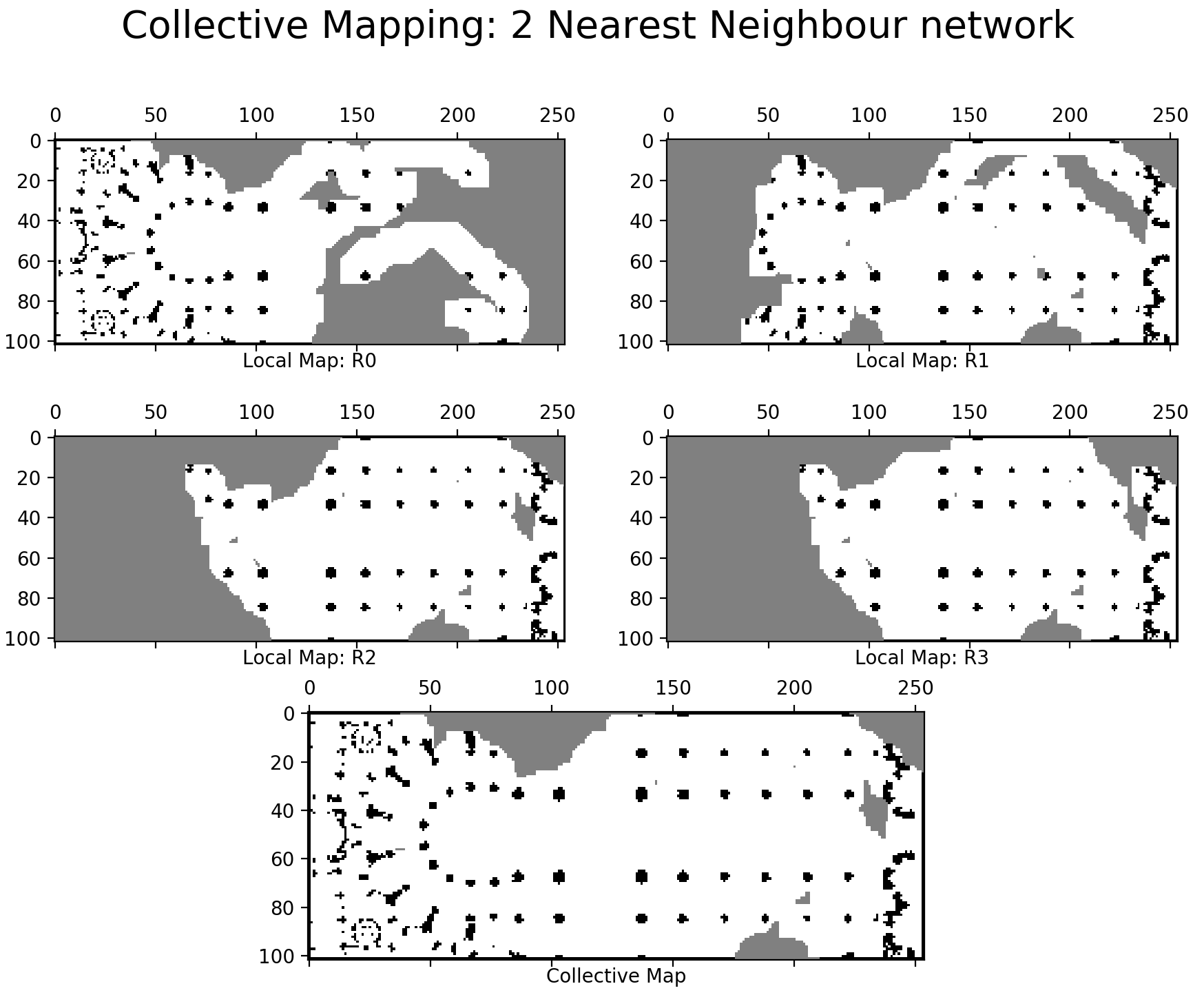}
\caption{\textsf{Examples of the different local maps computed by each robot on a given run when interacting with their $k=2$ closest neighbors.}}
\label{churchill}
\end{figure}

In the various simulation runs, a swarm of robots is deployed within a Basilica, whose floor layout is shown in Fig.~\ref{church}, and is tasked to map the interior of the building in a collective manner.
As the robots explore the environment, each unit builds its own partial map, see Fig.~\ref{churchill}.
The termination criterion for the simulation is that the union of these maps covers a 100~\% of the environment. Note that this termination criterion does require the full assembled map but in practice the swarm of robots shall continue mapping to detect possible dynamic changes in the layout. Hence, the termination criterion is only used here to assess the effectiveness of our proposed collective mapping approach.

\subsubsection{Scalability}

To test scalability, we run simulations with  $k=1$ nearest neighbor network topology and change the number of robots from 5, 10, 15, to 20.
The termination criterion is that the collective map is 100\% completed.
The results are shown in Table~\ref{tab:scalable}.
As the number of robots increases, the collective mapping duration reduces.
From 15 robots to 20 robots, the decrease in number of iterations is not as significant due to the limiting factor of the environment space.

\subsubsection{Robustness}
\begin{figure}[htp]
\centering
\includegraphics[width=0.48\textwidth]{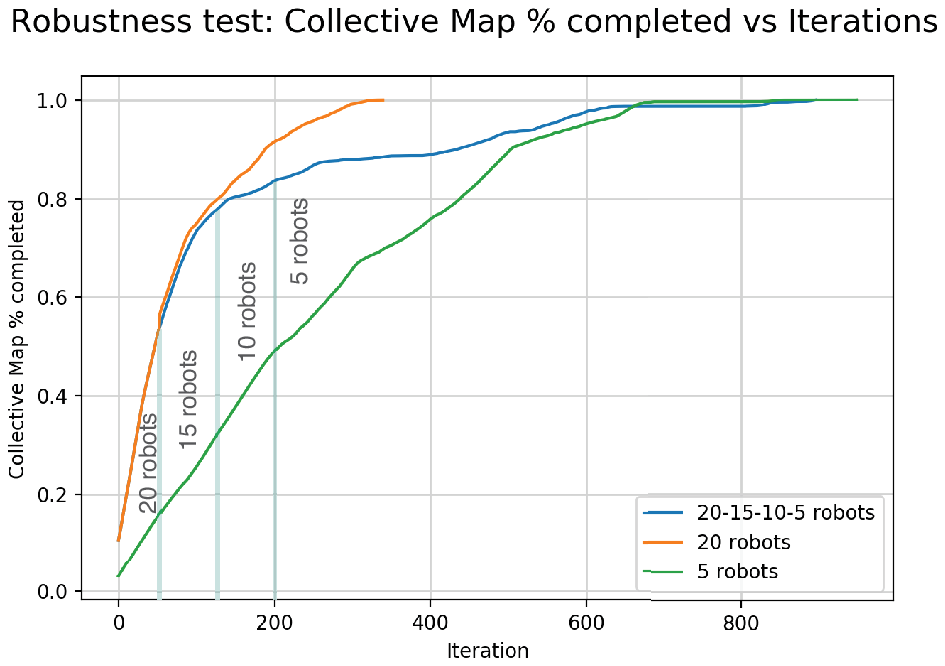}
\caption{\textsf{Robustness test result showing the completion of Collective Mapping despite a disrupted network(20-15-10-5 robots graph)---i.e. removing 5 agents every 50 iterations. }}
\label{fig:ds}
\end{figure}

To test robustness, we run simulations with $k=2$ nearest neighbor network topology.
%Figure~\ref{fig:ds} shows the collective map completion throughout the iterations.
Figure~\ref{fig:ds} shows the evolution of map coverage with time for three cases: a swarm of $20$ units, a swarm of $5$ units, and a swarm that starts with $20$ and where 5 robots are regularly removed from it every 50 iterations.
%There are three plots in this figure that show results of the simulation for 20 robots, 5 robots, and 20 starting with 5 robots dropped every 50 iterations until 5. The robots completed the task at 339, 948 and 896 iterations respectively.
The 20-15-10-5 robots graph shows that our collective mapping approach is robust, even when robots are dropping out of the network during the operation, the system is able to complete 100\% of the mapping.
Interestingly, we can see that the 20 robots graph and 20-15-10-5 robots graph separated right after 5 robots are removed at iteration 50.
The new 15 robots system started to slow down in its operation, and even more after subsequent removals. Initially, the 20-15-10-5 system operates well above the 5 robots system, after all the removals, both the system converges and completed the collective mapping 50 iterations apart.

\subsubsection{Network Effect}
In Table~\ref{tab:networkeffect} one can see the average number of iterations taken to fulfill this criterion for different number of nearest neighbors.
The simulations are based on 15 robots connected via the different network topology.
Chain is a static network where each robot is connected to two other particular robots.
0NN means there is no interaction between robots at all with ``NN" standing for ``nearest neighbors".
From 1NN to 6NN, these are the $k>0$ nearest neighbor dynamic networks.
The results in Table~\ref{tab:networkeffect} shows the drastic improvement from a 0NN to $(k>0)$NN, from a totally independent system to a swarming system.
From the results, it is observed that the dynamic network are more efficient than a static network in performing collective mapping, and it becomes more efficient as $k$ is increased.
Based on this initial study, 5NN seems to be the optimal network for this operation.
However, extensive study can be carried out to find the optimal network as in\cite{mateo2018optimal}.

\begin{table}[htp]
    \centering
    \begin{tabular}{|l|c|c|c|c|}
	\hline
	No. of robots & 5 & 10 & 15 & 20\\
	\hline
    No. of Iterations & 1273 & 982 & 583 & 562\\
    \hline
\end{tabular}
    \caption{Scalability test of Collective Mapping: Number of iterations needed for the collective to map 100\% of the floor.}
    \label{tab:scalable}
\end{table}

\begin{table}[htp]
    \centering
    \begin{tabular}{|p{1.01cm}|p{0.5cm}|p{0.5cm}|p{0.5cm}|p{0.5cm}|p{0.5cm}|p{0.5cm}|p{0.5cm}|p{0.5cm}|}
	\hline
	Network & Chain & 0NN & 1NN & 2NN & 3NN & 4NN & 5NN & 6NN\\
	\hline
    No. of Iterations & 585 & 1092 & 583 & 469 & 344 & 322 & 303 & 310\\
    \hline
\end{tabular}
    \caption{Effect of network on Collective Mapping}
    \label{tab:networkeffect}
\end{table}

\newpage
\section{Discussion}
Applying swarming concepts to robotic mapping operations opens interesting possibilities for the collective mapping of very large areas with a large number of robotic units.
The intrinsic properties of swarming systems suits the intended principles of multi-robot SLAM. This motivated us to adapt the robotic mapping from the swarm robotics perspective.
We applied a collective operation and computation framework to the mapping operation to define the collective mapping based on a decentralized process requiring only local computations combined with a cooperative control strategy, also defined locally for each unit.
This collective mapping concept is thoroughly assessed and tested with simulations.
The results show that this approach provides a decentralized, robust, and scalable mapping framework, beyond what is currently possible with Edge computing.
As a next step, we will consider investigating the flexibility of the swarming system during this collective mapping process, and study its capacity to map changing environments.
% The next step is to be able to test the collective mapping on an existing robotic platform developed by our research group.

In designing multi-robot systems, a great deal of attention is paid to the dynamics of the agents with relatively less attention focused on the effects of dynamic networks.
In most cases, the infrastructure is so as to guarantee a static network that provides a constant and reliable information flow to a central station.
However, it is known that different kinds of connectivities and degrees distributions can affect drastically the performance of the system.
In particular, it has been shown that limiting the number of interacting agents increases the system's responsiveness \cite{mateo17:_effec_correl_swarm_collec_respon,mateo2018optimal}.

%maybe change a name for fault-tolerant robust collective computing framework
Distributed heavy computations on swarming system is of particular interest in the area of environmental sensing. To this aim, we are currently developing a fault-tolerant robust collective computing framework suited for multi-robot systems and with fully decentralized operations in dynamic environments.
As the system is meant to operate in the physical world, and is equipped with advanced sensors for environmental data collection, it can perform reconstruction or prediction of a quantity of interest depending on the collective task at hand.
Such high-level applications tend to be complex, and thus require significant computational resources, which can be a deterring factor for many existing multi-robot system.
Yet, this novel capability would have significant implications for the collective operation and system adaptability to changing circumstances and dynamics environments.
It is worth adding that the proposed robust collective computing framework under development, given its distributed nature, should preserve robustness and flexibility. Scalability will be dependent on the effectiveness of the distributed network of communication. We reckon that this is the next advancement for multi-agent systems, also paving the way for AI-based operations, e.g. using collective reinforcement learning~\cite{panait2005cooperative}.

From the hardware perspective, all computations are performed by single-board computers placed in each agent, and in the absence of any central computer or any other supporting infrastructure. The system is expected to be capable of continuing and completing its robust collective computation with the removal or addition of any number of nodes. As a proof of concept, we successfully implemented and tested this robust collective computing framework on a swarm of buoys, developed by our group\cite{zoss17:_distr_system_auton_buoys_scalab}, connected by means of a low-bandwidth dynamic mesh network.
The presented work on Collective computing is therefore the cornerstone to this robust collective computing framework. Both are designed to be platform- and environment-agnostic, i.e. they are completely independent of (i) the robotic platform's hardware, (ii) the environmental equation that the system aims to solve, and (iii) the numerical method considered for the solution of the partial differential equations governing the environmental model.

\section{Conclusion}

We have presented a collective computing framework applied to the complex task of collective, distributed mapping.
In contrast with current approaches, this framework allows for a completely decentralized and distributed mapping, thus affording scalability to autonomous systems operating in large environments.
Simulations of this framework show the effectiveness, scalability, and robustness of the cooperative strategy where agents interact through a dynamic, switching network topology with fixed degree.

%\bibliographystyle{plain}
%\bibliography{references.bib}

\end{document}